\documentclass[10pt,twocolumn,letterpaper]{article}

\usepackage{cvpr}
\usepackage{enumitem}

\usepackage[dvipsnames]{xcolor}
\usepackage{enumitem} 
\usepackage{tabularx}
\usepackage{subcaption}
\usepackage{mathtools}
\usepackage{nccmath}
\usepackage{algorithm}
\usepackage{algorithmic}
\usepackage{array,multirow}
\usepackage{indentfirst}
\usepackage[accsupp]{axessibility} 

\DeclarePairedDelimiter{\nint}\lfloor\rceil

\DeclareMathOperator*{\argmin}{arg\,min}

\definecolor{cvprblue}{rgb}{0.21,0.49,0.74}
\usepackage[pagebackref,breaklinks,colorlinks,citecolor=cvprblue]{hyperref}

\def\paperID{11772}
\def\confName{CVPR}
\def\confYear{2024}

\title{Instance-Aware Group Quantization for Vision Transformers}
\author{Jaehyeon Moon$^{1, 2}$ \quad\quad\quad Dohyung Kim$^{1}$ \quad\quad\quad Junyong Cheon$^{1}$  \quad\quad\quad Bumsub Ham$^{1}$\thanks{Corresponding author.}\vspace*{0.2mm}\\
$^{1}$Yonsei University\quad\quad\quad$^{2}$Articron\\
\vspace{1mm}
{\url{https://cvlab.yonsei.ac.kr/projects/IGQ-ViT/}}}

\begin{document}
\maketitle

\begin{abstract}
    Post-training quantization (PTQ) is an efficient model compression technique that quantizes a pretrained full-precision model using only a small calibration set of unlabeled samples without retraining. PTQ methods for convolutional neural networks (CNNs) provide quantization results comparable to full-precision counterparts. Directly applying them to vision transformers (ViTs), however, incurs severe performance degradation, mainly due to the differences in architectures between CNNs and ViTs. In particular, the distribution of activations for each channel vary drastically according to input instances, making PTQ methods for CNNs inappropriate for ViTs. To address this, we introduce instance-aware group quantization for ViTs (IGQ-ViT). To this end, we propose to split the channels of activation maps into multiple groups dynamically for each input instance, such that activations within each group share similar statistical properties. We also extend our scheme to quantize softmax attentions across tokens. In addition, the number of groups for each layer is adjusted to minimize the discrepancies between predictions from quantized and full-precision models, under a bit-operation (BOP) constraint. We show extensive experimental results on image classification, object detection, and instance segmentation, with various transformer architectures, demonstrating the effectiveness of our approach.
\end{abstract}
\vspace{-4mm}
\section{Introduction}
\label{sec:intro}

Transformers~\cite{vaswani2017attention} can capture long-range dependencies across sequential inputs, which is of central importance in natural language processing, aggregating contextual information and providing discriminative feature representations. Recently, vision transformers~(ViTs)~\cite{dosovitskiy2020image} has demonstrated the effectiveness of transformers for images, providing state-of-the-art results on various visual recognition tasks, including image classification~\cite{liu2021swin, touvron2021training}, object detection~\cite{liu2021swin, zhang2021vit}, and semantic segmentation~\cite{liu2021swin, strudel2021segmenter, xie2021segformer}. However, a series of fully-connected~(FC) and self-attention layers in ViTs requires a substantial amount of memory and computational cost, making it challenging to deploy them on devices with limited resources~(\eg,~drones and mobile phones). The growing demand for ViTs to operate on the resource-constrained devices has led to increased interest in developing network quantization techniques for ViTs.

\begin{figure}[t]
\begin{center}
    \begin{subfigure}[b]{0.48\linewidth}
        \centering
        \includegraphics[width=1\linewidth]{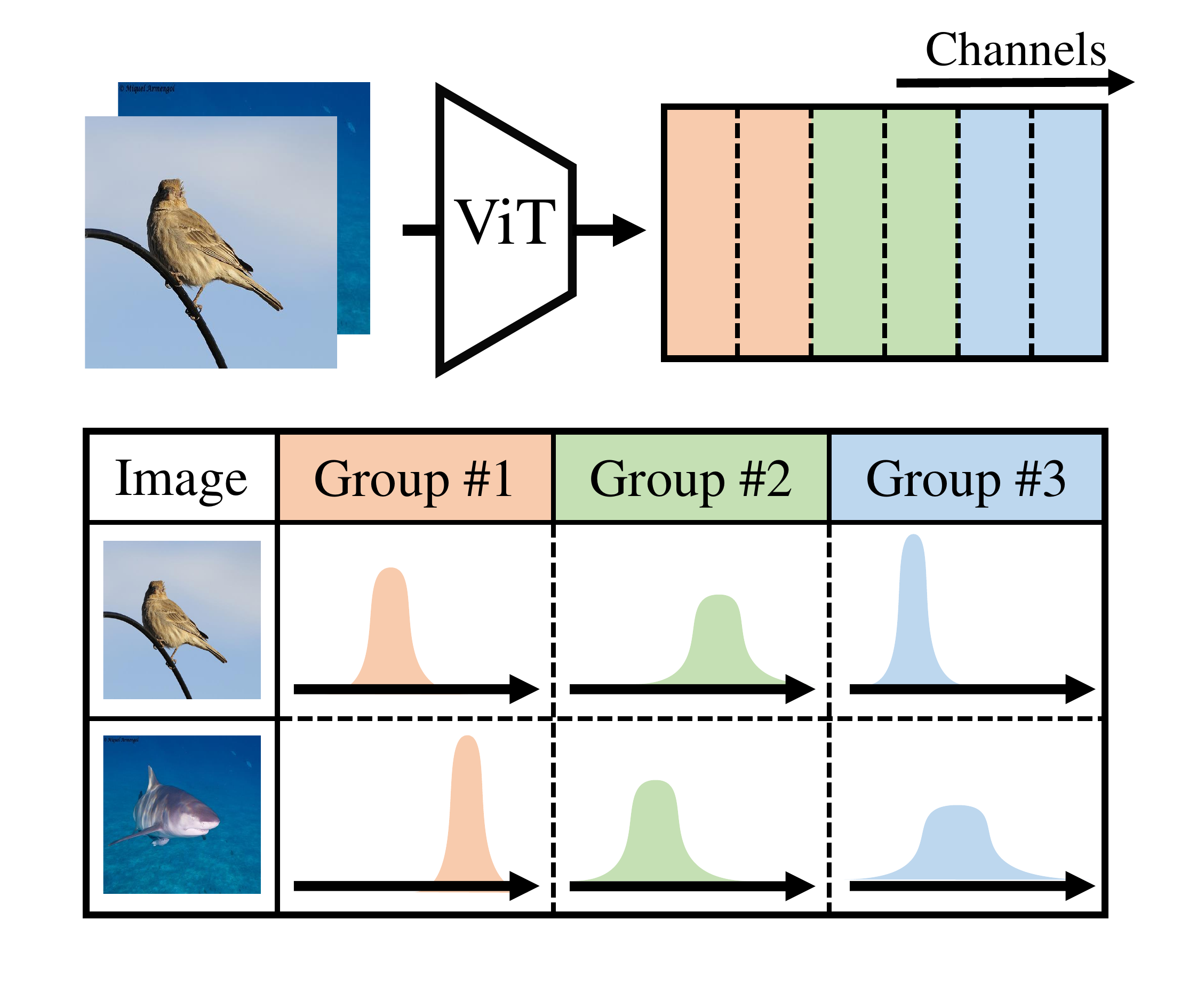}
        \captionsetup{font={small}}
        \caption{Group quantization}
        \label{fig:teaser_groupwise}
    \end{subfigure}
    \hspace{1mm}
    \begin{subfigure}[b]{0.48\linewidth}
        \centering
        \includegraphics[width=1\linewidth]{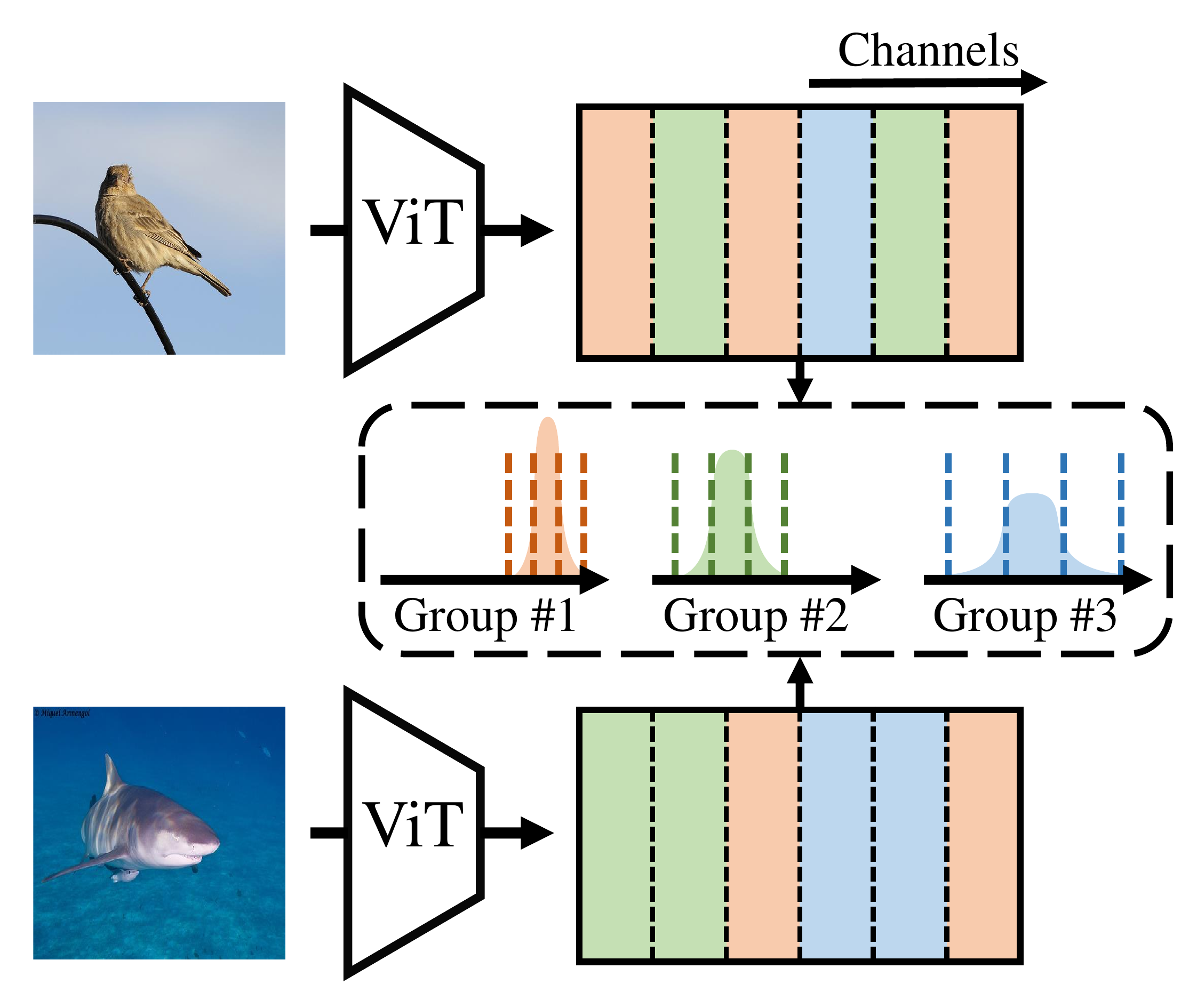}
        \captionsetup{font={small}}
        \caption{IGQ-ViT}
        \label{fig:teaser_ours}
    \end{subfigure}
    \vspace{-5mm}
    \captionsetup{font={small}}
    \caption{Visual comparison of group quantization and IGQ-ViT. (a) Conventional group quantization techniques~\cite{dai2021vs, shen2020q} divide consecutive channels uniformly into a number of groups without considering their dynamic ranges. The distribution of activations in each group varies significantly for individual input instances. (b) To alleviate this problem, IGQ-ViT proposes an instance-aware grouping technique that splits the channels of activation maps and softmax attentions across tokens dynamically for each input instance at runtime.}
    \label{fig:teaser}
\end{center}
\vspace{-7mm}
\end{figure}

Network quantization generally reduces bit-widths of weights and activations of a model for an efficient inference process, which can be categorized into two groups: Quantization-aware training (QAT) and post-training quantization (PTQ). QAT methods~\cite{esser2019learned, zhou2017incremental, zhuang2018towards} train full-precision models, while simulating the quantization process by inserting discretizers into networks to quantize, such that the discrepancy between the full-precision and quantized models is minimized in terms of accuracy. This suggests that QAT methods require entire training samples, and they are computationally expensive, making them impractical for the prompt deployment of neural networks. PTQ methods~\cite{li2021brecq, nagel2020up, wei2022qdrop}, on the other hand, calibrate quantization parameters~(\eg, quantization intervals, zero-points) from pretrained full-precision models, enabling faster quantization of networks compared to QAT methods with only a limited number of training samples (usually less than 1k).

Several PTQ methods for transformers~\cite{ding2022towards, liu2021post, yuan2021ptq4vit} apply layer-wise quantization techniques, where a single quantizer is applied to all activation values for efficiency. These methods, however, are not directly applicable for quantizing models using extremely low bit-widths~(\eg, 4-bit), due to the significant scale variation on the activations for each channel. Exploiting channel-wise quantizers~(\ie, applying different quantizers for each channel) could be a potential solution, but at the expense of computational overheads, due to floating-point summations of channel-wise outputs for matrix multiplication. Group quantization techniques~\cite{dai2021vs, shen2020q} could be an alternative to address this problem, where they divide consecutive channels uniformly into multiple groups, and apply a single quantizer for each group (Fig.~\ref{fig:teaser_groupwise}). However, we have observed that the channel-wise distributions of activation values vary largely among different samples, making conventional approaches inappropriate for ViTs.

In this paper, we present instance-aware group quantization for ViTs~(IGQ-ViT), that effectively and efficiently addresses the variations of channel-wise distributions across different input instances (Fig.~\ref{fig:teaser_ours}). Specifically, we split the channels of activation maps into multiple groups dynamically, such that the activation values within each group share similar statistical properties, and then quantize the activations within the group using identical quantization parameters. We also propose to use the instance-aware grouping technique to softmax attentions, since the distributions of attention values vary significantly according to tokens. In addition, we present a simple yet effective method to optimize the number of groups for individual layers, under a bit-operation (BOP) constraint. IGQ-ViT can be applied to various components in ViTs, including input activations of FC layers and softmax attentions, unlike previous methods~\cite{li2022repq, lin2021fq, liu2021post, yuan2021ptq4vit} that are limited to specific parts of transformer architectures. We demonstrate the effectiveness and efficiency of IGQ-ViT for various transformers, including ViT~\cite{dosovitskiy2020image} and its variants~\cite{liu2021swin, touvron2021training}, and show that IGQ-ViT achieves state-of-the-art results on standard benchmarks. We summarize the main contributions of our work as follows:
\begin{itemize}[leftmargin=*]
\item[$\bullet$] We introduce a novel PTQ method for ViTs, dubbed IGQ-ViT, that splits channels of activation maps into a number of groups dynamically according to input instances. We also propose to use the instance-aware grouping technique to split softmax attentions across tokens.
\item[$\bullet$] We present a group size allocation technique searching for an optimal number of groups for each layer given a BOP constraint.
\item[$\bullet$] We set a new state of the art on image classification~\cite{deng2009imagenet}, object detection, and instance segmentation~\cite{lin2014microsoft}, with various ViT architectures~\cite{dosovitskiy2020image, liu2021swin, touvron2021training}.
\end{itemize}
\vspace{-4mm}
\section{Related work}
\label{sec:related_work}

\paragraph{Network quantization.}
Network quantization aims at reducing bit-widths of weights and activations of neural networks. QAT methods simulate the quantization process by applying a round function to weights and activations of the network. Since derivatives of the round function is either zero or infinite, they approximate the gradients~(\eg, using the straight-through estimator~\cite{bengio2013estimating}) to train the network with backpropagation. These methods also adjust the derivatives of the round function~\cite{kim2021distance, lee2021network} or train quantization parameters jointly with network weights based on task losses~\cite{esser2019learned, jung2019learning}. For better convergence of the training process, many heuristics have been introduced, \eg, progressively shrinking bit-widths~\cite{zhuang2018towards} or freezing parts of the network weights~\cite{park2020profit, zhou2017incremental}. Quantized networks using QAT show performance comparable to or even better then full-precision counterparts. However, the quantization process is computationally demanding, requiring a significant amount of training time. PTQ offers an alternative approach to quantizing neural networks. Instead of training full-precision models and simulating the quantization process at training time, PTQ methods calibrate quantization parameters~(\eg,~quantization intervals) using a subset of training samples. Early efforts focus on optimizing the quantization parameters to minimize the difference between floating-point and quantized values~\cite{banner2019post, nahshan2021loss}. Another line of research proposes to consider distributions of weights and/or activations to design quantizers. For instance, the work of~\cite{fang2020post} has observed that network weights follow a bell-shaped distribution. Based on this, it introduces piecewise linear quantizers that assign different quantization intervals according to the magnitudes of activations, performing better compared to uniform quantizers. Recent PTQ methods learn to either round up or down network weights by using a reconstruction error of layer outputs~\cite{nagel2020up} or exploiting the Hessian of training losses~\cite{li2021brecq}, and they have proven the effectiveness on CNN architectures~(\eg, ResNet~\cite{he2016deep}, MobileNetV2~\cite{sandler2018mobilenetv2}).

\vspace{-5mm}
\paragraph{Transformer quantization.}
While ViTs~\cite{dosovitskiy2020image} and the variants~\cite{liu2021swin, touvron2021training} have become increasingly popular in computer vision, the unique structure and characteristics of ViT architectures makes network quantization challenging. For example, PTQ methods for CNNs~\cite{banner2019post, li2021brecq, nagel2020up, nahshan2021loss} do not perform well on quantizing softmax attentions and GELU activations in transformers, suggesting that directly applying them for ViT quantization results in significant performance degradation~\cite{liu2021post}. To date, only a limited number of PTQ methods have been developed for ViTs. The work of~\cite{liu2021post} estimates quantization parameters that maximize similarities between full-precision and quantized outputs of linear operations, and proposes to preserve a relative order of attention values after quantization. APQ-ViT~\cite{ding2022towards} introduces a calibration metric to minimize the discrepancies between full-precision and quantized outputs, while maintaining the power-law distribution of softmax attentions. PTQ4ViT~\cite{yuan2021ptq4vit} introduces twin uniform quantizers to handle asymmetric distributions in softmax attentions and GELU activations effectively. Most PTQ methods for ViTs exploit a single quantizer for all channels, suggesting that they do not consider the distributions of activation values across channels, typically having extreme scale variations. Recent works~\cite{lin2021fq, li2022repq} attempt to alleviate the scale variation problem efficiently. FQ-ViT~\cite{lin2021fq} proposes to consider inter-channel scale variations for LayerNorm~\cite{ba2016layer}, and exploits channel-wise quantizers with the constraint of the ratio of quantization intervals being power-of-two values. This enables using bit-shift operations, calculating mean and variance of LayerNorm in an integer level. The scale reparameterization technique, introduced by RepQ-ViT~\cite{li2022repq}, allows to use layer-wise quantizers, instead of adopting channel-wise ones, by adjusting the affine factors of LayerNorm and the weights of FC layers. However, this technique applies to the activations for LayerNorm only, and does not fully address the inter-channel scale variations for other layers in transformers.

Similar to ours, the works of~\cite{bondarenko2021understanding, dai2021vs, shen2020q, wang2022quantformer} adopt group quantization techniques for transformers. For instance, Q-bert~\cite{shen2020q} and VS-quant~\cite{dai2021vs} divide consecutive channels uniformly into a number of groups without considering the dynamic range of each channel, and thus the channels assigned to each group do not follow similar distributions. PEG~\cite{bondarenko2021understanding} alleviates this issue by sorting the activations across channels~\wrt the dynamic ranges during calibration, before grouping the channels. Quantformer~\cite{wang2022quantformer} proposes to use a differentiable search~\cite{cai2020rethinking, liu2018darts} for QAT in order to group channels of activation maps. The channels assigned to particular groups are however fixed after calibrating pre-trained networks for PTQ in the group quantization techniques~\cite{bondarenko2021understanding, dai2021vs, shen2020q}, which makes them inappropriate for ViTs having diverse channel distributions according to input instances. In contrast, our approach apply group quantization along channels of activation maps and tokens of softmax attentions dynamically at runtime for each input instance, without additional parameters for PTQ.
\vspace{-2mm}
\section{Method}
\label{sec:method}
In this section, we provide a brief description of uniform quantizer~(Sec.~\ref{sec:3_1}). We then present our approach in detail, including IGQ-ViT~(Sec.~\ref{sec:3_2}) and a group size allocation technique~(Sec.~\ref{sec:3_3}).
\subsection{Uniform quantizer}
\label{sec:3_1}
Given a floating-point value $x$ and the quantization bit-width $b$, uniform quantizers discretize the inputs into a finite set of values with equally spaced intervals. To this end, it normalizes the floating-point value $x$ using a scale parameter $s$, calibrate the normalized value with a zero-point~$z$, and clip the output as follows: 
\begin{equation}
   \label{eq:quant_uniform}
   \hat{x} = \text{clip}(\nint{\frac{x}{s}} + z, 0, 2^{b} - 1),
\end{equation}
where the scale parameter~$s$ and zero-point~$z$ are defined as:
\begin{equation}
   \label{eq:quant_params}
   s = \frac{u - l}{2^{b} - 1},\quad z = \text{clip}(\nint{-\frac{l}{s}}, 0, 2^{b} - 1).
\end{equation}
We denote by $u$ and $l$ upper and lower bounds of the quantizer, respectively. $\nint{.}$ is a rounding function, and $\text{clip}(., m, n)$ restricts an input to the range with lower and upper bounds of $m$ and $n$, respectively. The quantized output is then obtained as follows:
\begin{equation}
   \label{eq:dequant_uniform}
   Q(x; s, z) = s(\hat{x} - z).
\end{equation}

\begin{figure}[t]
   \begin{center}
   \begin{subfigure}{0.162\textwidth}
      \centering
      \includegraphics[width=1\linewidth]{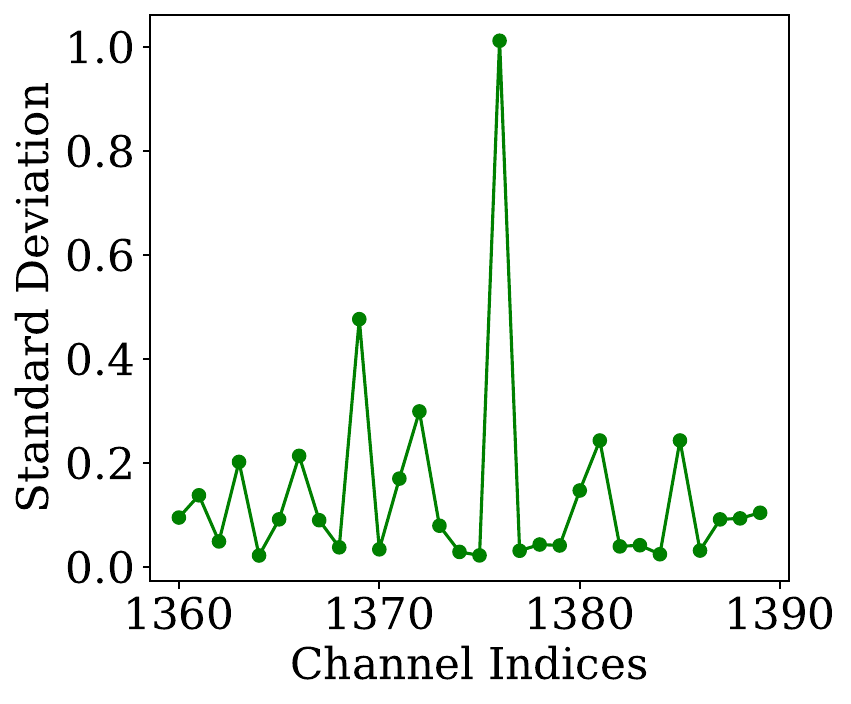}
      \captionsetup{font={normalsize}}
      \caption{}
      \label{fig:observation_b}
   \end{subfigure}
   \begin{subfigure}{0.152\textwidth}
      \centering
      \includegraphics[width=1\linewidth]{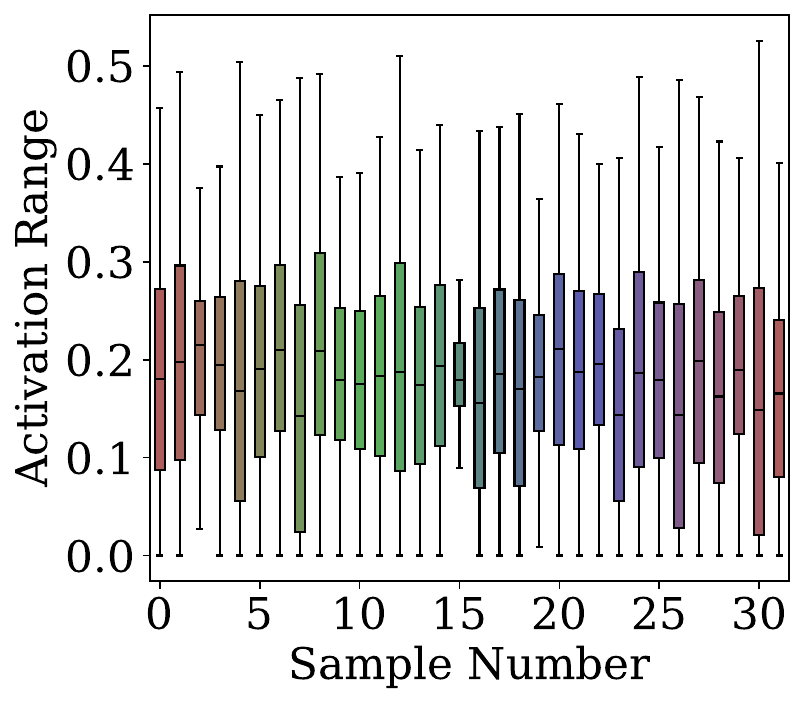}
      \captionsetup{font={normalsize}}
      \caption{}
      \label{fig:observation_c}
   \end{subfigure}
   \begin{subfigure}{0.152\textwidth}
      \centering
      \includegraphics[width=1\linewidth]{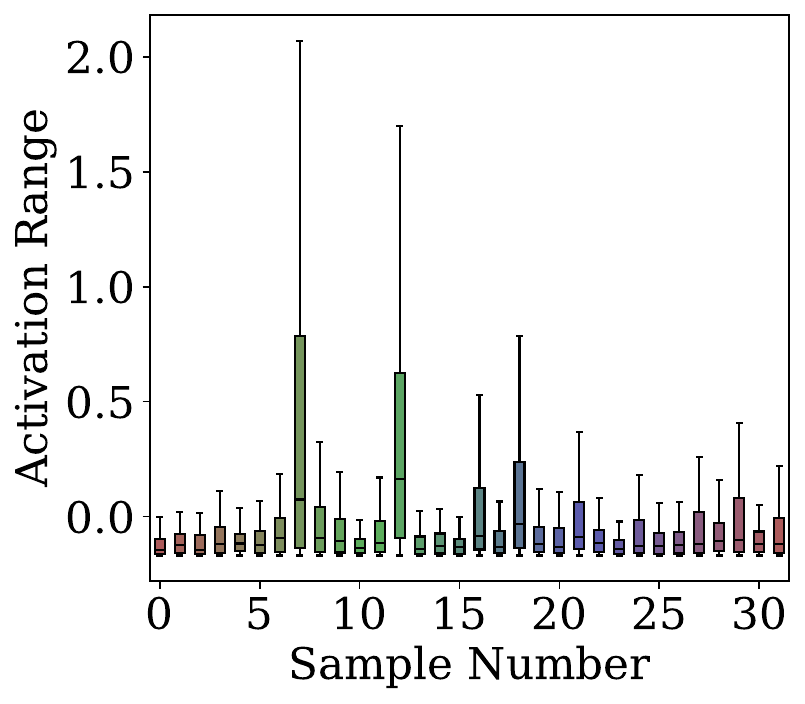}
      \captionsetup{font={normalsize}}
      \caption{}
      \label{fig:observation_d}
  \end{subfigure}
  \vspace{-2mm}
  \captionsetup{font={small}}
  \caption{(a) Plots of standard deviations of activations across channels for DeiT-S~\cite{touvron2021training}; (b-c) Boxplots of activation values across different input instances for a particular channel of ResNet-50~\cite{he2016deep} and DeiT-S, respectively. We use ImageNet~\cite{deng2009imagenet} for the visualizations. We have observed that there is a significant scale variation across channels, and the activation ranges for each channel change drastically among different samples for ViTs, in contrast to CNNs.}
  \label{fig:observation}
\end{center}
\vspace{-8mm}
\end{figure}

\vspace{-2mm}
\subsection{IGQ-ViT}
\label{sec:3_2}
Following the work of~\cite{liu2021post}, we quantize all network weights except for the positional embedding. We also quantize input activations of FC layers in the multi-layer perceptron~(MLP) block, and the activations for the multi-head self-attention~(MSA) block, including queries, keys, values, and softmax attentions. We exploit uniform quantizers for all weights and activations in ViTs.

\vspace{-3mm}
\subsubsection{IGQ for linear operations}
\label{sec:method_sqlinear}
In the following, we first provide empirical observations on input activations of FC layers, and explain the details of our IGQ framework for linear operations.

\vspace{-4mm}
\paragraph{Distributions of activations across channels.}
Most quantization frameworks~\cite{banner2019post, li2021brecq, nagel2020up, liu2021post} exploit layer-wise quantizers for activations, applying a single quantization parameter for all channels for efficient inference. However, we have observed that the input activations of FC layers have significant scale variations across channels (Fig.~\ref{fig:observation}(a)). Similar findings can be found in~\cite{li2022repq, lin2021fq}. This suggests that layer-wise quantizers degrade the quantization performance significantly, as they cannot handle scale variations across different channels. Although adopting separate quantizers for individual channels could be an effective strategy for overcoming the scale variation problem, this requires a summation of a floating-point output for every channel, which is computationally expensive. We have also found that the ranges of these activations for each channel vary drastically among different input instances~(Fig.~\ref{fig:observation}(b, c)), since ViTs do not have preceding BatchNorm~\cite{ioffe2015batch} layers in contrast to state-of-the-art CNNs~(\eg, ResNet~\cite{he2016deep}, MobileNetV2~\cite{sandler2018mobilenetv2}). Conventional approaches~(\eg, \cite{bondarenko2021understanding, dai2021vs, shen2020q, wang2022quantformer}) exploit a fixed quantization interval~(\ie, from lower to upper bounds of the quantizer) for every input instance, thus cannot adapt to such diverse distributions across different samples.

\vspace{-4mm}
\paragraph{Instance-aware grouping across channels.}
We introduce an instance-aware group quantization framework for linear operations that alleviates the scale variation problem, while maintaining efficiency. We split the channels of activation maps into $G_1$ groups based on statistical properties, where activation values within each group are quantized with identical quantization parameters. We assign the channels of activations to appropriate groups, and optimize the scale parameter~${s}_{i}$ and the zero-point~${z}_{i}$ for the $i$-th group. Specifically, given floating-point activations of ${\bf{X}} \in \mathbb{R}^{N \times C}$ and a set of candidate quantizers $\{Q_{i}\}_{i=1}^{G_1}$, where we denote by $N$ and $C$ as the number of tokens and channels, respectively, we define a distance metric between the $c$-th channel of the activations ${\bf{X}}$, denoted by~${\bf{X}}_{c}$, and the quantizer $Q_{i}$ as follows:
\begin{equation}
   \label{eq:distance}
   d({\bf{X}}_{c}, Q_{i}) = (\min({\bf{X}}_{c}) - {u}_{i})^{2} + (\max({\bf{X}}_{c}) - {l}_{i})^{2}
\end{equation}
where ${u}_{i}$ and ${l}_{i}$ are upper and lower bounds of the quantizer $Q_{i}$, respectively. We then assign each channel of the activation ${\bf{X}}$ to one of candidate quantizers with the minimum distance as follows:

\begin{equation}
   \label{eq:assignment}
   \pi(c) = \argmin_{i} d({\bf{X}}_{c}, Q_{i})
\end{equation}
where we denote by $\pi(c)$ a group index assigned to the $c$-th channel. The upper and lower bounds of quantizers are then optimized by minimizing the distances as follows:
\begin{equation}
   \label{eq:calibration}
   {u}^{\ast}_{i}, {l}^{\ast}_{i} = \argmin_{{u}_{i}, {l}_{i}} \sum_{\pi(c)=i} d({\bf{X}}_{c}, Q_{i}).
\end{equation}
We optimize ${{u}}_{i}$ and ${{l}}_{i}$ by solving Eq.~\eqref{eq:assignment} and Eq.~\eqref{eq:calibration} alternately similar to the expectation-maximization (EM) algorithm, which guarantees the convergence~\cite{wu1983convergence}. Finally, we obtain quantization parameters of each group~(\ie, ${\bf{s}}$ and ${\bf{z}}$) using Eq.~\eqref{eq:quant_params}. At test time, we fix the quantization parameters, and assign the channels to appropriate groups using Eq.~\eqref{eq:assignment}.

 \begin{table}[t] 
   \setlength{\tabcolsep}{0.3em}
   \small
   \captionsetup{font={small}}
   \centering
   \bgroup
   \setlength{\tabcolsep}{0.5pt}
   \caption{Comparison of BOPs for a 4-bit DeiT-B~\cite{touvron2021training} model using various quantization strategies. We denote by `Model' the required BOP for layer-wise quantization. In contrast to layer-wise quantization, IGQ-ViT involves additional computations, including (1) computing the min/max values of each channel, (2) assigning channels to quantizers with the minimum distance, and (3) summing the outputs of each group in a floating-point format. The corresponding BOPs for these steps are denoted by `Minmax', `Assign', and `FP sum', respectively.}
   \vspace{-2mm}
   \label{tab:bitops}
   \begin{tabular}{l|cccc|c}
   \hline
   \multicolumn{1}{c|}{\bf{Methods}} & \bf{Model} & \bf{Minmax} & \bf{Assign} & \bf{FP sum} & \bf{Total} \\ 
   \hline
   Layer-wise & 340.3G & - & - & - & 340.3G \\ 
   \hline
   IGQ-ViT(\#groups=4) & 340.3G & 0.99G & 0.57G & 1.57G & 343.4G \\
   IGQ-ViT(\#groups=8) & 340.3G & 0.99G & 1.14G & 3.66G & 346.1G \\
   IGQ-ViT(\#groups=16) & 340.3G & 0.99G & 2.28G & 7.84G & 351.4G \\
   \hline
   \end{tabular}
   \egroup
   \vspace{-2mm}
\end{table}

\vspace{-6mm}
\paragraph{Computational overhead.}
\label{sec:computational_overhead}
Compared to layer-wise quantization, our approach requires (1) computing the min/max values of each channel, (2) assigning channels to quantizers with the minimum distance, and (3) summing the floating-point outputs of each group. Specifically, consider a matrix multiplication between a quantized activation $Q({\bf{X}})$ and the quantized weight $Q({\bf{W}})$ with a group size of $G_1$. The quantized activation $Q({\bf{X}})$ is obtained by partitioning ${\bf{X}}$ into a number of groups across channels using Eq.~\eqref{eq:assignment},~\ie ${\bf{X}} = [{\bf{X}}_{1}, ..., {\bf{X}}_{G_1}]$, followed by quantizing ${\bf{X}}_{i}$ with a scale parameter of $s_{i}$, where $i \in \{1, ..., G_1\}$. The matrix multiplication between $Q({\bf{X}})$ and $Q({\bf{W}})$ can then be represented as follows:
\begin{equation}
   \label{eq:group_wise_quant}
   Q({\bf{X}})Q({\bf{W}}) = s^{w} \cdot (\sum_{i=1}^{G_1} {s}_{i} \cdot \hat{\bf{X}}_{i}\hat{\bf{W}}_{i}).
\end{equation}
Note that we omit zero-points for clarity. We denote by $s^{w}$ the scale parameter for ${\bf{W}}$. $\hat{\bf{X}}_{i}$ and $\hat{\bf{W}}_{i}$ are obtained by applying Eq.~\eqref{eq:quant_uniform} to ${\bf{X}}_{i}$ and ${\bf{W}}_{i}$, the channels and rows of $\bf{X}$ and $\bf{W}$ associated with group $i$, respectively. Computing Eq.~\eqref{eq:group_wise_quant} requires the summation of floating-point matrices for each group~(\ie, ${s}_{i} \cdot \hat{\bf{X}}_{i}\hat{\bf{W}}_{i}$), which can be reduced with sufficiently small values of $G_1$. As the values of $G_1$, we use no more than 16 in our experiments, which is extremely small compared to the number of channels, usually scaling up to over a thousand. We show in Table~\ref{tab:bitops} BOPs of IGQ-ViT for DeiT-B~\cite{touvron2021training} quantized with 4-bit. We can see that IGQ-ViT introduces only 3.3\% additional BOPs for a group size of 16, compared to layer-wise quantization.

\begin{figure}[t]
   \captionsetup{font={small}}
   \begin{center}
       \includegraphics[width=0.9\linewidth]{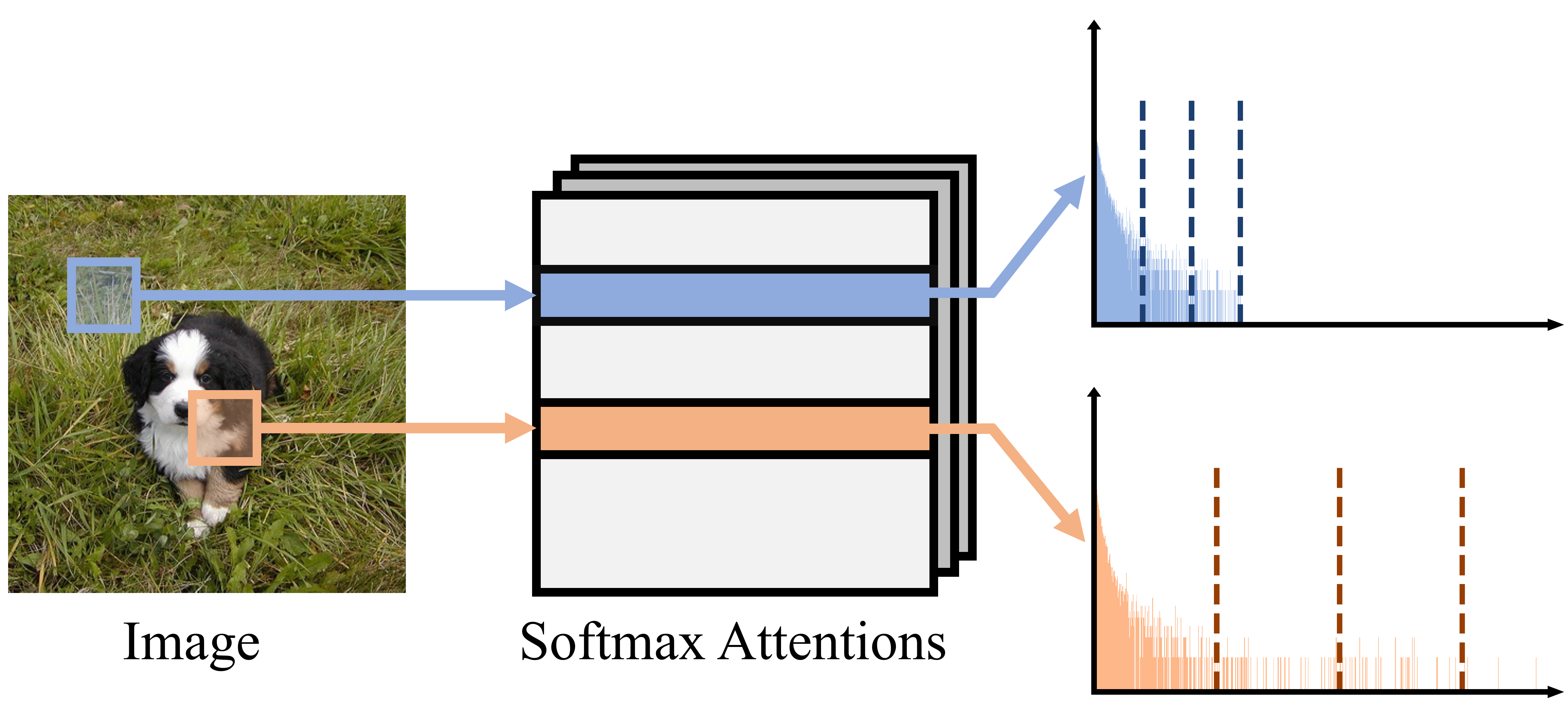}
   \end{center}
   \vspace{-5mm}
   \caption{Distributions of softmax attentions across tokens. We can see that the distributions are different significantly across tokens. Our approach can handle this issue by splitting the rows of softmax attentions into several groups and applying separate quantizers for each group, such that the attentions assigned to each group share similar statistical properties.}
   \label{fig:attn}
   \vspace{-3mm}
\end{figure}

\begin{figure}[t]
   \begin{center}
   \begin{subfigure}{0.23\textwidth}
      \centering
      \includegraphics[width=1\linewidth]{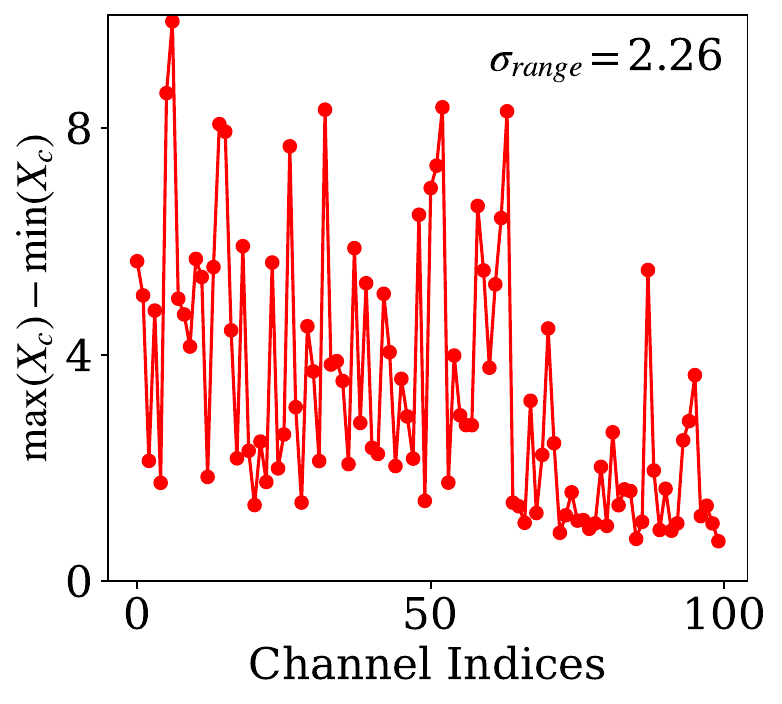}
      \label{fig:gsa_a}
   \end{subfigure}
   \begin{subfigure}{0.23\textwidth}
      \centering
      \includegraphics[width=1\linewidth]{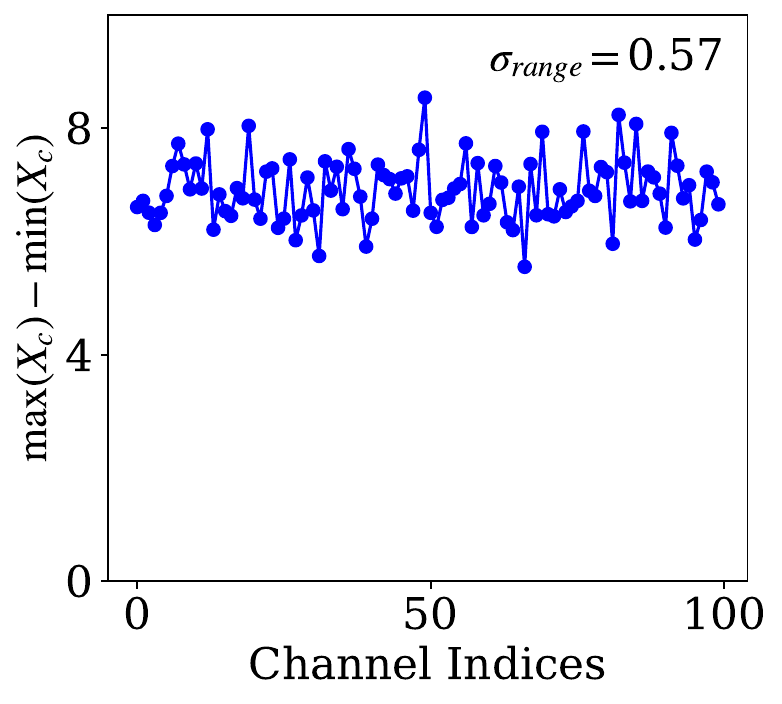}
      \label{fig:gsa_b}
  \end{subfigure}
  \vspace{-8mm}
  \captionsetup{font={small}}
  \caption{Comparisons for dynamic ranges of activation values across channels, chosen from different layers of ViT-S~\cite{dosovitskiy2020image}. $\sigma_{range}$ is the standard deviation of the dynamic ranges of channels for each layer. We can see that the degree of scale variations across channels varies according to the layer, suggesting that the number of groups for each layer would be adjusted.}
  \label{fig:gsa}
\end{center}
\vspace{-8mm}
\end{figure}

\vspace{-3mm}
\subsubsection{IGQ for softmax attentions}
\label{sec:method_sqattention}

Here, we present our observation for the distribution of softmax attentions, and present the details of IGQ for softmax attentions.

\vspace{-4mm}
\paragraph{Distributions of softmax attentions.}
ViTs capture correlations between tokens through softmax attentions. The distribution of attention values varies drastically across different tokens (Fig.~\ref{fig:attn}). Therefore, using a single quantization parameter to quantize softmax attentions degrades quantization performance severely. Separate quantizers might be exploited for individual tokens to handle the attention values, but this requires a large number of quantizers, and needs to adjust the quantization parameters for each instance.

\vspace{-4mm}
\paragraph{Instance-aware grouping across tokens.}
We extend our approach in order to quantize softmax attentions across tokens. We split softmax attentions for tokens~(\ie, the rows of the softmax attentions) into $G_2$ groups, according to the distribution of attention values. Specifically, given softmax attentions ${\bf{A}} \in \mathbb{R}^{H \times N \times N}$ and a set of quantizers $\{Q_{j}\}_{j=1}^{G_2}$, where we denote by $H$ the number of heads, we define the distance between each row of softmax attentions and a quantizer $Q_{j}$ as follows:
\begin{equation}
   \label{eq:distance_soft}
   d({\bf{A}}_{n}, Q_{j}) = (\max({\bf{A}}_{n}) - v_{j})^{2}
\end{equation}

where we denote by ${\bf{A}}_{n}$ and ${v}_{j}$ the $n$-th row of ${\bf{A}}$ and the upper bound of the quantizer $Q_{j}$, respectively. Note that we set lower bounds to 0, as all attention values are positive. We then optimize ${v}_{j}$ with the EM algorithm and set the quantization parameters for quantizer $Q_{j}$ using Eq.~\eqref{eq:quant_params}.

\begin{table*}[t] 
   \setlength{\tabcolsep}{0.3em}
   \small
   \captionsetup{font={small}}
   \centering
   \bgroup
   \def\arraystretch{1.15}
   \renewcommand{\arraystretch}{1.2}
   \setlength{\tabcolsep}{2.5pt}
   \caption{Quantitative results of quantizing ViT architectures on ImageNet~\cite{deng2009imagenet}. W/A represents the bit-width of weights (W) and activations (A), respectively. We report the top-1 validation accuracy (\%) with different group sizes for comparison. The numbers of other quantization methods are taken from~\cite{ding2022towards, li2022repq}. $^\dagger$: Results without using a group size allocation~(\ie, a fixed group size for all layers).
   }
   \label{tab:imagenet}
   \vspace{-0.7cm}
   \begin{tabular}{l c c c c c c c c c c}\\
   \hline
   \multicolumn{1}{c}{\bf{Method}} & \bf{\#bits(W/A)} & \bf{ViT-T} & \bf{ViT-S} & \bf{ViT-B} & \bf{DeiT-T} & \bf{DeiT-S} & \bf{DeiT-B} & \bf{Swin-T} & \bf{Swin-S} & \bf{Swin-B}\\  
   \hline
   Full-precision  & 32/32 & 75.47 & 81.39 & 84.54 & 72.21 & 79.85 & 81.80 & 81.39 & 83.23 & 85.27 \\
   \hline
   PTQ4ViT~\cite{yuan2021ptq4vit}  & 4/4 & 17.45 & 42.57 & 30.69 & 36.96 & 34.08 & 64.39 & - & 76.09 & 74.02 \\
   APQ-ViT~\cite{ding2022towards}  & 4/4 & 17.56 & 47.95 & 41.41 & 47.94 & 43.55 & 67.48 & - & 77.15 & 76.48 \\
   RepQ-ViT~\cite{li2022repq}  & 4/4 & - & 65.05 & 68.48 & 57.43 & 69.03 & 75.61 & - & 79.45 & 78.32 \\
   \hline
   IGQ-ViT$^\dagger$~(\#groups=8)  & 4/4 & \underline{54.46} & \underline{72.99} & \underline{77.91} & \underline{61.94} & \underline{74.01} & \underline{78.85} & \underline{76.91} & \underline{80.54} & \underline{82.88} \\
   IGQ-ViT$^\dagger$~(\#groups=12)  & 4/4 & \bf{55.66} & \bf{73.46} & \bf{79.13} & \bf{62.28} & \bf{74.52} & \bf{79.21} & \bf{77.53} & \bf{80.79} & \bf{83.08} \\
   \hline
   IGQ-ViT~(\#groups=8)  & 4/4 & \underline{55.29} & \underline{73.18} & \underline{78.28} & \underline{62.25} & \underline{74.23} & \underline{79.04} & \underline{77.19} & \underline{80.66} & \underline{83.02} \\
   IGQ-ViT~(\#groups=12)  & 4/4 & \bf{56.01} & \bf{73.61} & \bf{79.32} & \bf{62.45} & \bf{74.66} & \bf{79.23} & \bf{77.77} & \bf{80.98} & \bf{83.14} \\
   IGQ-ViT (upper bound)  & 4/4 & 57.59 & 74.73 & 79.88 & 62.55 & 75.08 & 79.74 & 78.58 & 81.48 & 83.53 \\
   \hline
   PTQ4ViT~\cite{yuan2021ptq4vit}  & 6/6 & 64.46 & 78.63 & 81.65 & 69.68 & 76.28 & 80.25 & - & 82.38 & 84.01 \\
   APQ-ViT~\cite{ding2022towards}  & 6/6 & 69.55 & 79.10 & 82.21 & 70.49 & 77.76 & 80.42 & - & 82.67 & 84.18 \\
   RepQ-ViT~\cite{li2022repq}  & 6/6 & - & 80.43 & \underline{83.62} & 70.76 & 78.90 & 81.27 & - & \underline{82.79} & 84.57 \\
   \hline
   IGQ-ViT$^\dagger$~(\#groups=8)  & 6/6 & \underline{72.90} & 80.07 & 83.11 & 70.71 & \underline{78.92} & \underline{81.34} & \underline{80.23} & 82.55 & 84.43 \\
   IGQ-ViT$^\dagger$~(\#groups=12)  & 6/6 & \bf{73.63} & \bf{80.66} & \bf{83.63} & \bf{71.02} & \bf{79.17} & \bf{81.48} & \bf{80.59} & 82.66 & \bf{84.70} \\
   \hline
   IGQ-ViT~(\#groups=8)  & 6/6 & \underline{73.19} & \underline{80.48} & 83.46 & \underline{70.92} & \underline{79.04} & \underline{81.44} & \underline{80.48} & 82.65 & \underline{84.62} \\
   IGQ-ViT~(\#groups=12)  & 6/6 & \bf{73.77} & \bf{80.76} & \bf{83.77} & \bf{71.15} & \bf{79.28} & \bf{81.71} & \bf{80.89} & \bf{82.86} & \bf{84.82} \\
   IGQ-ViT (upper bound)  & 6/6 & 74.61 & 80.99 & 84.27 & 71.42 & 79.42 & 81.75 & 81.20 & 83.08 & 85.06 \\
   \hline
   \end{tabular}
   \egroup
   \vspace{-5mm}
\end{table*}

\begin{figure}[t]
\begin{algorithm}[H]
   \caption{IGQ-ViT}
   \label{alg:algorithm}
   \small
   \begin{algorithmic}[1]
      \STATE \textbf{Hyperparameter}:
      Number of iterations $N_\text{iter}$; update period of group size $T$.
 
      \textbf{Input}:
      Pre-trained model; calibration set; target BOPs $N_{bop}$.
 
      \STATE For inputs of FC layers and softmax attentions, compute the distance between their channels/rows and quantizers using Eq.~\eqref{eq:distance} or Eq.~\eqref{eq:distance_soft}.
      \FOR{$k = 1, ..., N_\text{iter}$}
         \STATE Update the parameters for each group using Eq.~\eqref{eq:assignment} and Eq.~\eqref{eq:calibration}.
      \IF{$k\ \%\ T == 0$}
         \STATE Update the group size for each layer using Eq.~\eqref{eq:group_size_allocation}.
      \ENDIF
      \ENDFOR
      \STATE Obtain quantization parameters ${\bf{s}}$, ${\bf{z}}$ using Eq.~\eqref{eq:quant_params}.
 
      \STATE \textbf{Output}:
      Quantized model
   \end{algorithmic}
\end{algorithm}
\vspace{-2em}
\end{figure}

\subsection{Group size allocation}
\label{sec:3_3}
We observe that activations and softmax attentions in different layers show different amount of scale variations across channels and tokens, respectively, indicating that using the same number of groups for different layers might be suboptimal (Fig.~\ref{fig:gsa}). To address this, we search for the optimal group size for each layer that minimizes the discrepancy between the predictions from quantized and full-precision models, under a BOP constraint. However, the search space for finding the optimal group sizes is exponential~\wrt the number of layers $L$, which is intractable for a large model. We propose a group size allocation technique that efficiently optimizes the group size for each layer within such a large search space. Concretely, we define a perturbation metric for a particular layer $\psi(.)$ as the Kullback-Leibler (KL) divergence between predictions of the model before and after quantization as follows:
\begin{equation}
   \label{eq:group_size_metric}
   \psi(g, l) = D_\text{KL}(y_{l}||y_{l}^{g}),
\end{equation}
where we denote by $y_{l}$ and $y_{l}^{g}$ the predictions of the model before and after quantizing $l$-th layer with a group size of $g$, respectively. Note that we quantize all other layers except for the $l$-th layer for computing the predictions of $y_{l}$ and $y_{l}^{g}$, to account for the effects of quantization on different layers. For a target BOP $N_{bop}$, we formulate the group size allocation as a integer linear programming (ILP) problem, and search for the optimal group size for each layer, such that an overall perturbation of the model is minimized as follows:
\begin{equation}
   \label{eq:group_size_allocation}
   {\bf{g}}^{\ast} = \argmin_{\bf{g}} \sum_{l=1}^{L} \psi(g_l, l)\ \text{s.t.}\ B({\bf{g}}) \le N_{bop},
\end{equation}
where ${\bf{g}}=\{g_{l}\}_{l=1}^{L}$, and $g_{l}$ is the group size assigned to $l$-th layer. We denote by $B({\bf{g}})$ the BOP of the model with the group sizes of ${\bf{g}}$. We solve Eq.~\eqref{eq:group_size_allocation} with the PULP~\cite{roy2020pulp} library, using group size for each layer within the set of $\{4, 6, 8, 10, 12, 16\}$. We allocate the group sizes for every $T$ alternating steps of Eq.~\eqref{eq:assignment} and Eq.~\eqref{eq:calibration}, where $T$ is a hyperparameter. We show in Algorithm~\ref{alg:algorithm} an overall quantization process of our approach.
\section{Experiments}
\label{sec:experiments}
In this section, we describe our experimental settings (Sec.~\ref{sec:4_1}), and evaluate IGQ-ViT on image classification, object detection and semantic segmentation (Sec.~\ref{sec:4_2}). We then present a detailed analysis of our approach~(Sec.~\ref{sec:4_3}).

\begin{table*}[t] 
    \setlength{\tabcolsep}{0.3em}
    \small
    \centering
    \bgroup
    \def\arraystretch{1.15}
    \renewcommand{\arraystretch}{1.2}
    \setlength{\tabcolsep}{2.5pt}
    \caption{Quantitative results of quantizing Mask R-CNN~\cite{he2017mask} and Cascade Mask R-CNN~\cite{cai2018cascade} using Swin transformers~\cite{liu2021swin} on COCO~\cite{lin2014microsoft}. We report the box average precision $\text{AP}^\text{box}$ for object detection and the mask average precision $\text{AP}^\text{mask}$ for instance segmentation.}
    \vspace{-0.7cm}
    \label{tab:coco}
    \begin{tabular}{l c c c c c c c c c c c}\\
    \hline
        \multicolumn{1}{c}{\multirow{3}{*}{\bf{Method}}} & \multicolumn{1}{c}{\multirow{3}{*}{\bf{\#bits(W/A)}}} & \multicolumn{4}{c}{\bf{Mask R-CNN}} & \multicolumn{6}{c}{\bf{Cascade Mask R-CNN}} \\  
        \cline{3-12}
        & & \multicolumn{2}{c}{\bf{Swin-T}} & \multicolumn{2}{c}{\bf{Swin-S}} & \multicolumn{2}{c}{\bf{Swin-T}} & \multicolumn{2}{c}{\bf{Swin-S}} & \multicolumn{2}{c}{\bf{Swin-B}}\\  
        \cline{3-12}
        & & AP\textsuperscript{box} & AP\textsuperscript{mask} & AP\textsuperscript{box} & AP\textsuperscript{mask} & AP\textsuperscript{box} & AP\textsuperscript{mask} & AP\textsuperscript{box} & AP\textsuperscript{mask} & AP\textsuperscript{box} & AP\textsuperscript{mask}\\  
    \hline
    Full-precision  & 32/32 & 46.0 & 41.6 & 48.5 & 43.3 & 50.4 & 43.7 & 51.9 & 45.0 & 51.9 & 45.0 \\
    \hline
    PTQ4ViT~\cite{yuan2021ptq4vit}  & 4/4 & 6.9 & 7.0 & 26.7 & 26.6 & 14.7 & 13.5 & 0.5 & 0.5 & 10.6 & 9.3 \\
    APQ-ViT~\cite{ding2022towards}  & 4/4 & 23.7 & 22.6 & \underline{44.7} & 40.1 & 27.2 & 24.4 & 47.7 & 41.1 & 47.6 & \underline{41.5} \\
    RepQ-ViT~\cite{li2022repq}  & 4/4 & 36.1 & 36.0 & 44.2 & \underline{40.2} & 47.0 & 41.4 & \underline{49.3} & \underline{43.1} & - & - \\
    \hline
    IGQ-ViT (\#groups=8) & 4/4 & \underline{40.5} & \underline{38.5} & \underline{44.7} & \bf{41.3} & \underline{48.4} & \underline{42.3} & \bf{50.5} & \bf{44.0} & \underline{50.4} & \bf{44.0} \\
    IGQ-ViT (\#groups=12) & 4/4 & \bf{41.0} & \bf{38.8} & \bf{44.8} & \bf{41.3} & \bf{48.5} & \bf{42.4} & \bf{50.5} & \bf{44.0} & \bf{50.5} & \bf{44.0} \\
    \hline
    PTQ4ViT~\cite{yuan2021ptq4vit}  & 6/6 & 5.8 & 6.8 & 6.5 & 6.6 & 14.7 & 13.6 & 12.5 & 10.8 & 14.2 & 12.9 \\
    APQ-ViT~\cite{ding2022towards}  & 6/6 & \underline{45.4} & \underline{41.2} & 47.9 & 42.9 & 48.6 & 42.5 & 50.5 & 43.9 & \underline{50.1} & \underline{43.7} \\
    RepQ-ViT~\cite{li2022repq}  & 6/6 & 45.1 & \underline{41.2} & 47.8 & 43.0 & \underline{50.0} & 43.5 & \underline{51.4} & 44.6 & - & - \\
    \hline
    IGQ-ViT (\#groups=8) & 6/6 & \underline{45.4} & \bf{41.5} & \bf{48.2} & \underline{43.1} & \bf{50.4} & \underline{43.7} & \bf{51.9} & \underline{44.9} & \bf{51.9} & \bf{45.0} \\
    IGQ-ViT (\#groups=12) & 6/6 & \bf{45.5} & \bf{41.5} & \bf{48.2} & \bf{43.2} & \bf{50.4} & \bf{43.8} & \bf{51.9} & \bf{45.0} & \bf{51.9} & \bf{45.0} \\
    \hline
    \end{tabular}
    \egroup
    \vspace{-0.5cm}
\end{table*}

\subsection{Implementation details}
\label{sec:4_1}
We evaluate our IGQ-ViT framework on the tasks of image classification, object detection, and instance segmentation. We use the ImageNet~\cite{deng2009imagenet} dataset for image classification, which contains approximately 1.2M images for training, and 50K for validation. We use COCO~\cite{lin2014microsoft} for object detection and instance segmentation, which includes 118K training, 5K validation, and 20K test images. We adopt various transformer architectures, including ViT~\cite{dosovitskiy2020image}, DeiT~\cite{touvron2021training}, and Swin transformer~\cite{liu2021swin}, for image classification. For the tasks of object detection and instance segmentation, we use Mask R-CNN~\cite{he2017mask} and Cascade Mask R-CNN~\cite{cai2018cascade} with Swin transformers as the backbone. Following~\cite{ding2022towards, li2022repq}, we randomly sample 32 images from the ImageNet~\cite{deng2009imagenet} dataset for image classification, and a single image from COCO~\cite{lin2014microsoft} for object detection and instance segmentation to calibrate the quantization parameters. We apply our instance-aware grouping technique for all input activations of FC layers, and softmax attentions. More detailed settings are available in the supplement.

\begin{table}[t] 
    \setlength{\tabcolsep}{0.3em}
    \small
    \captionsetup{font={small}}
    \centering
    \bgroup
    \setlength{\tabcolsep}{2.5pt}
    \caption{Quantitative comparison of our instance-aware group quantization technique with various configurations under a 4/4-bit setting. We denote by `Linear' and `Attention' the quantization method for linear operations and softmax attentions, respectively. For applying our method, we use a group size of 8 for all layers.}
    \vspace{-5mm}
    \label{tab:ablation}
    \begin{tabular}{c c | c c c}\\
    \hline
    \multicolumn{1}{c}{\bf{Linear}} & \bf{Attention} & \bf{ViT-S} & \bf{DeiT-B} & \bf{Swin-T} \\  
    \hline
    Layer-wise & Layer-wise & 42.82 & 62.23 & 55.51 \\
    Layer-wise & Ours & 53.69 & 65.05 & 62.27 \\
    Ours & Layer-wise & 57.32 & 75.51 & 72.89 \\
    Ours & Ours & \bf{72.99} & \bf{78.85} & \bf{76.91} \\
    \hline
    Ours & Row-wise & 73.22 & 78.88 & 77.19 \\
    Channel-wise & Ours & 74.56 & 79.69 & 78.39 \\
    Channel-wise & Row-wise & \bf{74.73} & \bf{79.74} & \bf{78.58} \\
    \hline
    \end{tabular}
    \egroup
    \vspace{-4mm}
\end{table}

\subsection{Results}
\label{sec:4_2}
\paragraph{Results on ImageNet.}
We show in Table~\ref{tab:imagenet} the top-1 accuracy (\%) on the validation split of ImageNet~\cite{deng2009imagenet} with various ViT architectures. We report the accuracy with an average group size of 8 and 12. We summarize our findings as follows: (1)~Our IGQ-ViT framework with 8 groups already outperforms the state of the art except for ViT-B~\cite{dosovitskiy2020image} and Swin-S~\cite{liu2021swin} under 6/6-bit setting, while using more groups further boosts the performance. (2) Our approach under 4/4-bit setting consistently outperforms RepQ-ViT~\cite{li2022repq} by a large margin. Similar to ours, RepQ-ViT also addresses the scale variations between channels, but it can be applied to the activations with preceding LayerNorm only. In contrast, our method handles the scale variations on all input activations of FC layers and softmax attentions, providing better results. (3) Our group size allocation technique boosts the quantization performance for all models, indicating that using the same number of groups for all layers is suboptimal. (4) Exploiting 12 groups for our approach incurs less than 0.9\% accuracy drop, compared to the upper bound under the 6/6-bit setting. Note that the results of upper bound are obtained by using a separate quantizer for each channel of activations and each row of softmax attentions.

\vspace{-5mm}
\paragraph{Results on COCO.}
We show in Table~\ref{tab:coco} the quantization results for object detection and instance segmentation on COCO~\cite{lin2014microsoft}. We quantize the backbones of Swin transformers~\cite{liu2021swin} and the convolutional layers in the neck and head of Mask R-CNN~\cite{he2017mask} and Cascade Mask R-CNN~\cite{cai2018cascade}. We observe that PTQ4ViT~\cite{yuan2021ptq4vit} and APQ-ViT~\cite{ding2022towards}, that use layer-wise quantizers for activations, do not perform well. In contrast, IGQ-ViT outperforms the state of the art with 8 groups only, and the quantization performance further boosts by exploiting more groups. In particular, it provides the results nearly the same as the full-precision ones for the the 6/6-bit setting. This suggests that scale variations across different channels or tokens are much more critical for object detection and instance segmentation.

\subsection{Discussion}
\label{sec:4_3}

\begin{table}[t]
    \setlength{\tabcolsep}{0.3em}
    \small
    \captionsetup{font={small}}
    \centering
    \bgroup
    \setlength{\tabcolsep}{2.5pt}
    \caption{Quantitative comparison of quantizing transformer architectures using various group quantization techniques under a 4/4-bit setting, with a group size of 8 for all linear operations. Note that we use layer-wise quantization for softmax attentions for a fair comparison.}
    \label{tab:group_wise_quant}
    \centering
    \begin{tabular}{l | c c c}
    \hline
    \multicolumn{1}{c|}{\bf{Grouping methods}} & \bf{ViT-S} & \bf{DeiT-B} & \bf{Swin-T} \\
    \hline
    No grouping (\#groups=1) & 42.82 & 62.23 & 55.51 \\
    Grouping consecutive channels~\cite{dai2021vs, shen2020q} & 41.04 & 65.50 & 70.39 \\
    Sorting before grouping channels~\cite{bondarenko2021understanding} & 41.26 & 62.61 & 56.04 \\
    \hline
    Ours & \bf{57.32} & \bf{75.51} & \bf{72.89} \\
    \hline
    \end{tabular}
    \egroup
    \vspace{-7mm}
\end{table}

\paragraph{Comparison with different quantizers.}
We compare in Table~\ref{tab:ablation} the results of the variants of our method adopting different types of quantizers on input activations of FC layers and softmax attentions. From the first four rows, we can see that our approach outperforms layer-wise quantization by a large margin, both for linear operations and softmax attentions. This indicates that adopting a single quantization parameter for all channels and rows without considering their individual distributions can severely limit the quantization performance. The last three rows compare the results of our approach with channel/row-wise quantization. We observe that the difference in performance between our approach and channel/row-wise quantization is less than 1.8\% for three different models. With a small group size, our framework can achieve comparable performance to the upper bound, while maintaining efficiency.

Table~\ref{tab:group_wise_quant} shows the results of quantizing ViT architectures using various group quantization techniques, including~\cite{bondarenko2021understanding, dai2021vs, shen2020q}, and ours. While the works of~\cite{dai2021vs, shen2020q} divide consecutive channels uniformly into a number of groups, the method of~\cite{bondarenko2021understanding} first sorts channels~\wrt the dynamic ranges before partitioning them into groups. In contrast, we dynamically assign channels to groups according to the statistical properties of the channels. We find that our approach outperforms other methods by a large margin, indicating that fixing the channels assigned to each group can degrade the quantization performance significantly. We also observe that sorting the channels~\wrt their dynamic ranges during calibration does not boost the quantization performance for DeiT-B~\cite{touvron2021training} and Swin-T~\cite{liu2021swin}, suggesting that the dynamic range of each channel vary drastically across different input instances.

\begin{figure}[t]
\begin{center}
    \begin{subfigure}[b]{0.49\linewidth}
        \centering
        \includegraphics[width=1\linewidth]{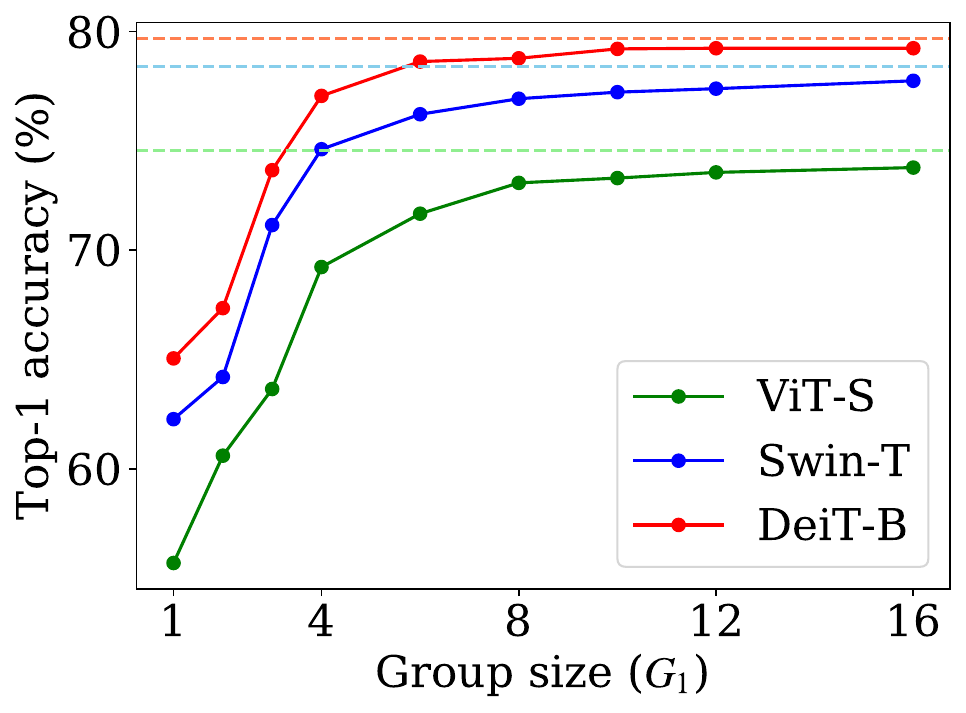}
        \label{fig:fc_grps}
    \end{subfigure}
    \begin{subfigure}[b]{0.49\linewidth}
        \centering
        \includegraphics[width=1\linewidth]{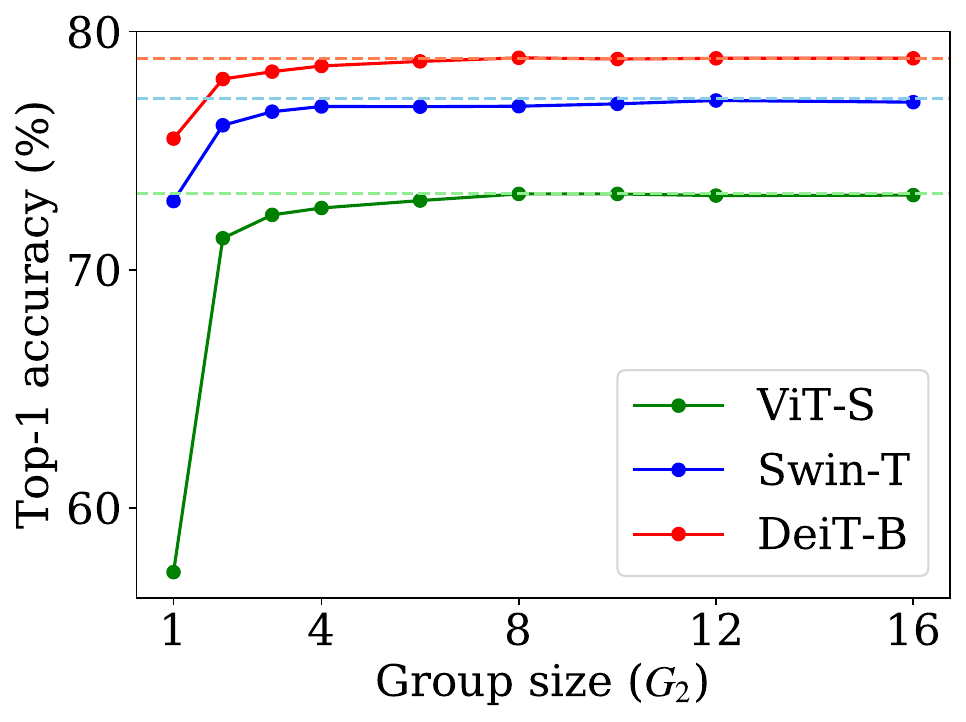}
        \label{fig:attn_grps}
    \end{subfigure}
    \vspace{-5mm}
    \captionsetup{font={small}}
    \caption{Top-1 validation accuracies on ImageNet~\cite{deng2009imagenet}~\wrt group sizes for linear operations ($G_1$, left) and that for softmax attentions ($G_2$, right). We set either $G_1$ or $G_2$ to 8, while varying the other to compute the accuracies. We report the quantization results of ViT-S~\cite{dosovitskiy2020image}, Swin-T~\cite{liu2021swin}, and DeiT-B~\cite{touvron2021training} under a 4/4-bit setting, with a fixed group size across different layers. We visualize the upper bounds with horizontal stripes of corresponding colors.}
    \label{fig:grps}
\end{center}
\vspace{-3mm}
\end{figure}

\begin{table}[t]
    \small
    \captionsetup{font={small}}
    \centering
    \bgroup
    \def\arraystretch{1.1}
    \setlength{\tabcolsep}{5pt}
    \caption{Quantitative comparison of ViT architectures with and without a group size allocation technique under the 4/4-bit setting.}
    \label{tab:group_size_allocation}
    \vspace{-2mm}
    \centering
    \begin{tabular}{c | c | c c c}
    \hline
    \multicolumn{1}{c|}{\bf{\#groups}} & \begin{tabular}{@{}c@{}}\bf{Group size}\\\bf{allocation}\end{tabular} & \bf{ViT-S} & \bf{DeiT-B} & \bf{Swin-T} \\ 
    \hline
    \multirow{2}{*}{6} &  & 71.44 & 78.44 & 76.31 \\
    & \checkmark & \bf{71.80} & \bf{78.57} & \bf{76.99} \\
    \hline
    \multirow{2}{*}{8} &  & 72.99 & 78.85 & 76.91 \\
    & \checkmark & \bf{73.18} & \bf{79.04} & \bf{77.19} \\
    \hline
    \multirow{2}{*}{10} &  & 73.34 & 79.00 & 77.21 \\
    & \checkmark & \bf{73.51} & \bf{79.19} & \bf{77.64} \\
    \hline
    \end{tabular}
    \egroup
    \vspace{-4mm}
\end{table}

\vspace{-4mm}
\paragraph{Analysis on group size.}
We show in Fig.~\ref{fig:grps} the results of IGQ-ViT according to the group size for linear operations (left) and softmax attentions (right). We can see that the quantization performance improves as the group size increases, for both linear operations and softmax attentions, demonstrating that using more groups better addresses the scale variation problem for channels and tokens. We also observe that the performance of our approach reaches near the upper bound with a small group size. This suggests that IGQ-ViT can effectively address the variations with a small amount of additional computations.

\begin{figure}[t]
\begin{center}
    \begin{subfigure}{0.23\textwidth}
        \centering
        \includegraphics[width=1\linewidth]{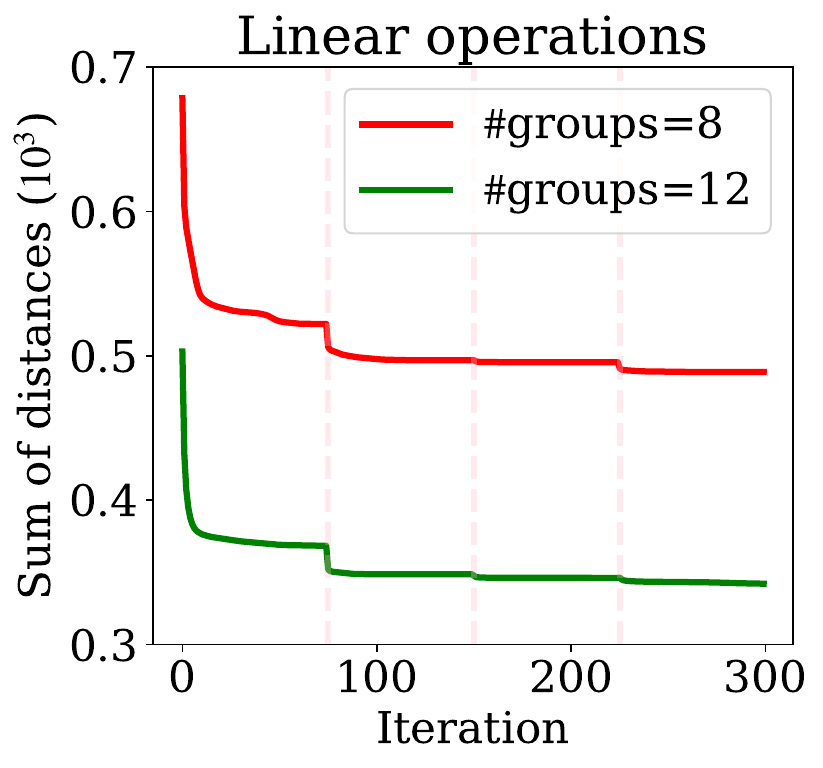}
        \label{fig:group_a}
        \vspace{-4mm}
    \end{subfigure}
    \begin{subfigure}{0.23\textwidth}
        \centering
        \includegraphics[width=1\linewidth]{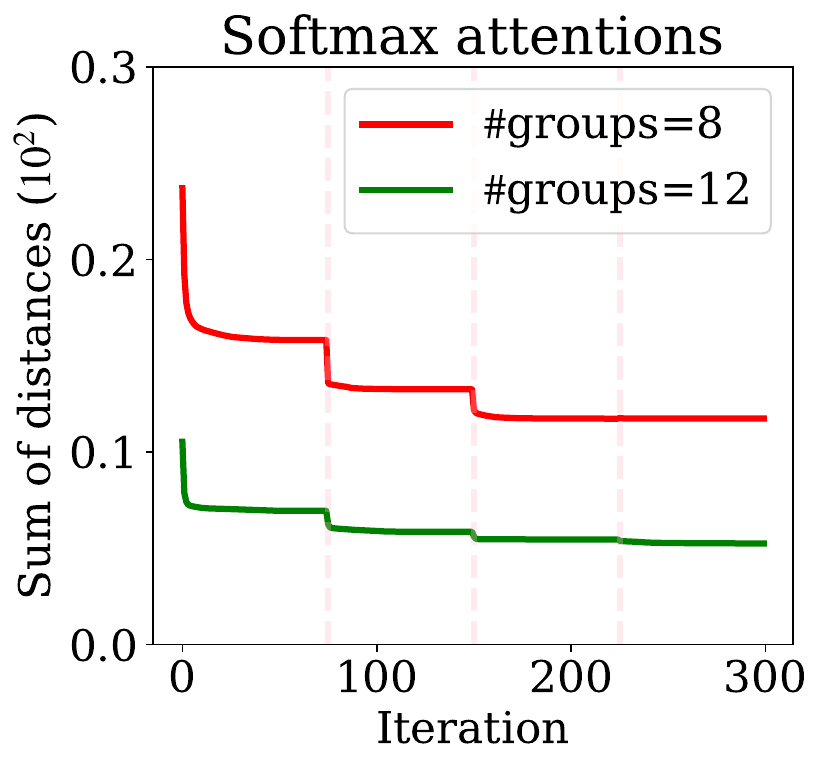}
        \label{fig:group_b}
        \vspace{-4mm}
    \end{subfigure}
    \begin{subfigure}{0.228\textwidth}
       \centering
       \includegraphics[width=1\linewidth]{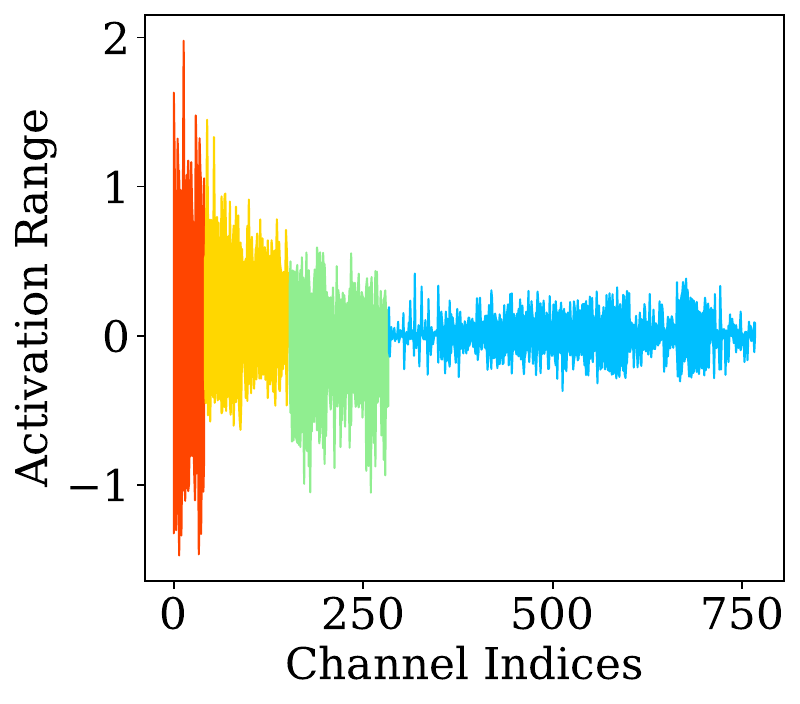}
       \label{fig:group_c}
    \end{subfigure}
    \begin{subfigure}{0.23\textwidth}
       \centering
       \includegraphics[width=1\linewidth]{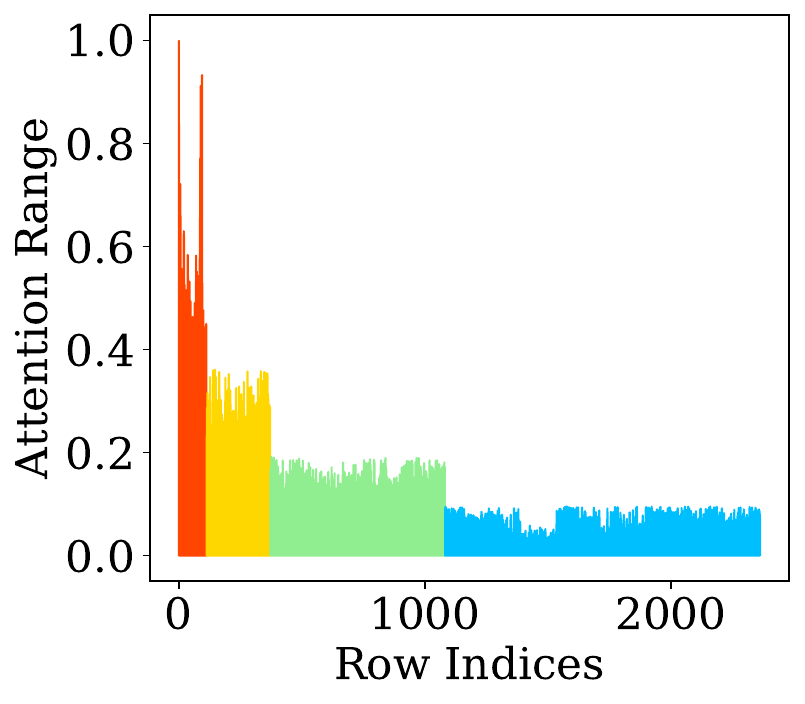}
       \label{fig:group_d}
   \end{subfigure}
   \vspace{-6mm}
   \captionsetup{font={small}}
   \caption{(Top)~Distances between channels of activations/rows of softmax attentions and quantizers in a particular layer of DeiT-S~\cite{touvron2021training}, that is, the distances as in Eq.~\eqref{eq:distance} and Eq.~\eqref{eq:distance_soft}, respectively; (Bottom)~Dynamic ranges of activations and attentions in a specific layer of DeiT-S~\wrt~the assigned group of the same color. Note that we sort the activations and attentions based on the group indices for the purpose of the visualization only.}
   \label{fig:group}
\end{center}
\vspace{-8mm}
\end{figure}

\vspace{-5mm}
\paragraph{Convergence analysis.}
We compare in Fig.~\ref{fig:group}(top) distances between channels of activation and quantizers in Eq.~\eqref{eq:distance} (rows of softmax attention and quantizers in Eq.~\eqref{eq:distance_soft}) over optimization steps. It shows that our algorithm converges quickly within a small number of optimization steps. We show in Fig.~\ref{fig:group}(bottom) the dynamic ranges of activations and attentions in a particular layer, along with their assigned groups after convergence. We can see that activations/attentions in each group share similar statistical properties, demonstrating that they can be effectively quantized with a single quantization parameter.

\vspace{-5mm}
\paragraph{Group size allocation.}
We compare in Table~\ref{tab:group_size_allocation} the results of our approach with/without the group size allocation technique. We can see that the group size allocation improves the quantization performance consistently, suggesting that assigning the same group size for all layers is suboptimal.

\vspace{-2mm}
\section{Conclusion}
\label{sec:conclusion}

We have observed that activations and softmax attentions in ViTs have significant scale variations for individual channels and tokens, respectively, across different input instances. Based on this, we have introduced a instance-aware group quantization framework for ViTs, IGQ-ViT, that alleviates the scale variation problem across channels and tokens. Specifically, our approach splits the activations and softmax attentions dynamically into multiple groups along the channels and tokens, such that each group shares similar statistical properties. It then applies separate quantizers for individual groups. Additionally, we have presented a simple yet effective method to assign a group size for each layer adaptively. We have shown that IGQ-ViT outperforms the state of the art, using a small number of groups, with various ViT-based architectures. We have also demonstrated the effectiveness of IGQ-ViT compared with its variants, including layer-wise quantizers, channel/row-wise quantizers, and state-of-the-art group quantizers, with a detailed analysis.

\vspace{-6mm}
\paragraph{Acknowledgements.}
This work was supported in part by the NRF and IITP grants funded by the Korea government (MSIT) (No.2023R1A2C2004306, No.RS-2022-00143524, Development of Fundamental Technology and Integrated Solution for Next-Generation Automatic Artificial Intelligence System, and No.2021-0-02068, Artificial Intelligence Innovation Hub).

{
    \small
    \bibliographystyle{ieeenat_fullname}
    \bibliography{main}
}

\clearpage
\maketitlesupplementary
\section*{A. More implementation details}
\label{sec:impl_details}

\subsection*{A.1. Weight quantization}
For weight quantization, we exploit a distinct quantizer for each output channel, following~\cite{li2022repq}. We designate the upper and lower bounds of weight quantizers with (100-$\epsilon$)-th and $\epsilon$-th percentiles of weight values in each output channel, where $\epsilon$ is a hyperparameter.

\subsection*{A.2. Hyperparameter settings}
We set $\epsilon$ as 5e-2 and 1e-3 for 4/4-bit and 6/6-bit settings, respectively. The values of $\epsilon$ are found using a grid search, such that KL divergence between predictions of full-precision and quantized ViT-T from calibration data is minimized. In addition, we set $N_\text{iter}$ and $T$ in Algorithm~1 as 300 and 75, respectively, which are large enough for our algorithm to be converged.

\subsection*{A.3. Perturbation metric for Mask R-CNN models}
We use the Mask R-CNN~\cite{he2017mask} and Cascade Mask R-CNN~\cite{cai2018cascade} with Swin transformer~\cite{liu2021swin} for the tasks of object detection and instance segmentation. The perturbation metric in Eq.~(9) cannot be applied directly to these models, typically having three outputs for each region of interest (RoI): (1) a class probability, (2) a bounding-box regression offset, and (3) a binary mask. The work of~\cite{he2017mask} proposes a multi-task loss on each sampled RoI as $L = L_{cls} + L_{box} + L_{mask}$, where the loss terms $L_{cls}$, $L_{box}$, and $L_{mask}$ are defined as the discrepancy between the outputs~(\ie, class probability, bounding-box offset, and binary mask) and corresponding ground-truth labels. Note that $L_{box}$ and $L_{mask}$ are defined only on the bounding-box and mask corresponding to the ground-truth class for each RoI, respectively. Refer to~\cite{he2017mask} for more details.

For Mask R-CNN and Cascade Mask R-CNN models, we modify the perturbation metric for each RoI in Eq.~(9) as follows:
\begin{equation}
    \label{eq:group_size_metric_mask_rcnn}
    \psi(g, l) = D_\text{KL}(y_{l}||y_{l}^{g}) + |b_{l} - b_{l}^{g}|+ |m_{l} - m_{l}^{g}|, \tag{A}
\end{equation}
where $y_{l}^{g}$, $b_{l}^{g}$, and $m_{l}^{g}$ are predictions of class probability, bounding-box offset, and binary mask, respectively, for a model in which the $l$-th layer is quantized with a group size of $g$. We denote by $y_{l}$, $b_{l}$, and $m_{l}$ the predictions of the model, where the $l$-th layer is left to be full-precision. Note that we consider only the distances for the ground-truth class during the computation of $|b_{l} - b_{l}^{g}|$ and $|m_{l} - m_{l}^{g}|$, following~\cite{he2017mask}. We use Eq.~\eqref{eq:group_size_metric_mask_rcnn} to solve for Eq.~(10) in the main paper.

\renewcommand\thefigure{A}
\begin{figure}[t]
    \captionsetup{font={small}}
    \begin{center}
        \includegraphics[width=\linewidth]{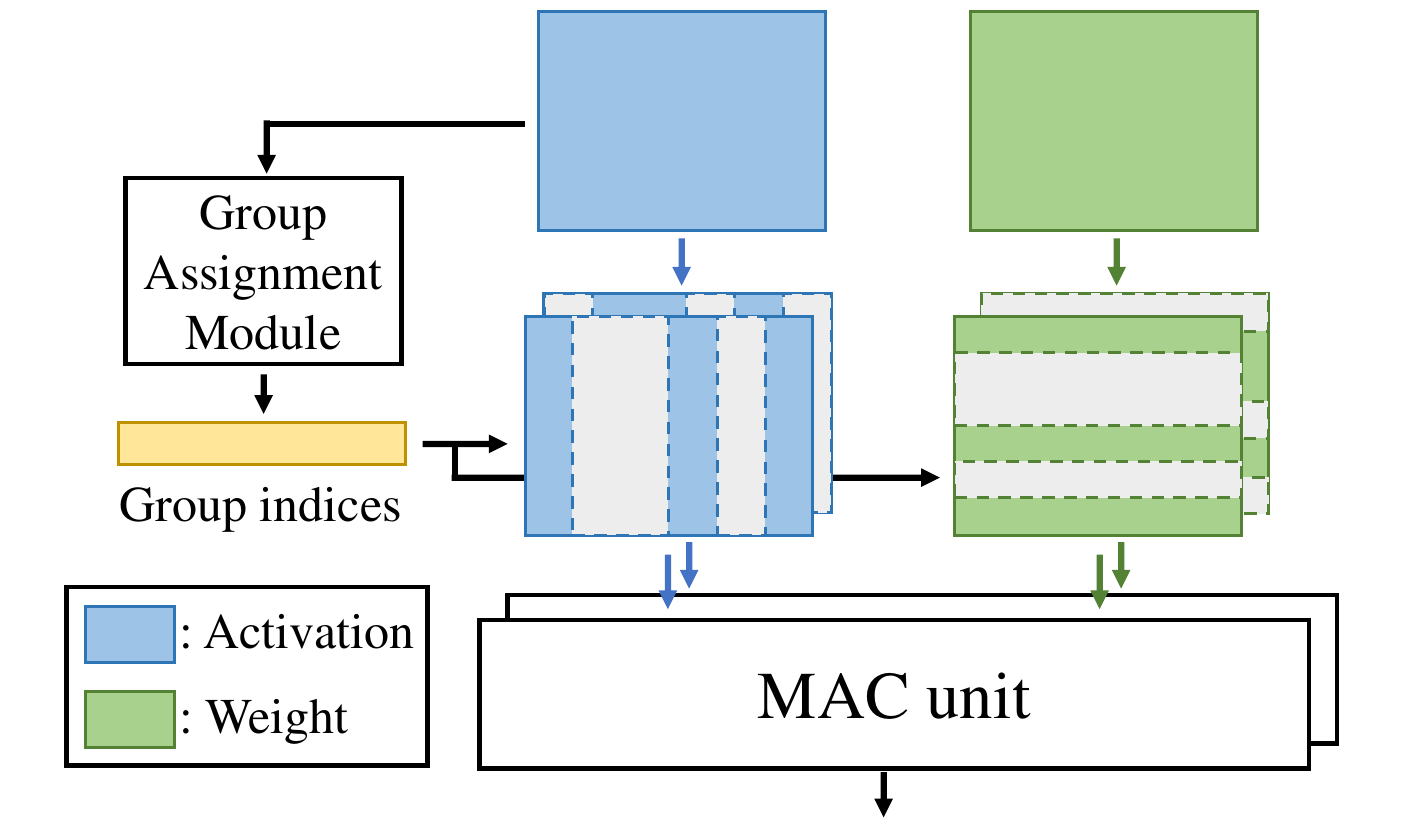}
    \end{center}
    \vspace{-5mm}
    \caption{Our instance-aware group quantization framework could be implemented using existing DNN accelerators, with a slight modification from the implementation of~\cite{dai2021vs}. One might leverage a group assignment module that computes the group indices for each channel using Eq.~(5), and the resulting indices are used to select channels that belong to each group during matrix multiplication. Corresponding rows of weight buffers are also selected.}
    \label{fig:supp}
    \vspace{-4mm}
\end{figure}

\section*{B. Compatibility with existing hardwares}
We believe that IGQ-ViT could be efficiently implemented using existing neural network accelerators, with a slight modification from the implementation of VS-quant~\cite{dai2021vs}. Specifically,~\cite{dai2021vs} divides channels of activations into a number of groups, and activation values assigned to each group are processed with separate multiply-accumulate (MAC) units. The outputs from each MAC unit are then scaled with different quantization parameters. In contrast, IGQ-ViT dynamically splits channels according to their statistical properties for each input instance. Compared to~\cite{dai2021vs}, IGQ-ViT requires additional computations, which includes computing the min/max values of each channel, and assigning channels to quantizers with the minimum distance. To address this, one might leverage a group assignment module that computes the min/max values of each channel, followed by obtaining group indices using Eq.~(5) (Fig.~\ref{fig:supp}). The resulting indices are then used to select channels that belong to each group. Finally, a separate MAC unit is applied for activation values within each group, which contains a single quantization parameter. Note that computing the group indices for each channel is computationally cheap in terms of BOPs~(See Table~1 in the main paper), and using an indexing scheme for efficient computation is a common practice in real devices. For example, the work of~\cite{yue202115} implements a module providing indices to dynamically detect sparsity patterns of weight and activation values in each group. It then uses the indices to skip groups of zero-valued weights and activations for efficiency (See Fig.~15.2.3 in~\cite{yue202115}).

\renewcommand\thefigure{B}
\begin{figure}[t]
  \begin{center}
      \begin{subfigure}[b]{0.45\linewidth}
          \centering
          \includegraphics[width=1\linewidth]{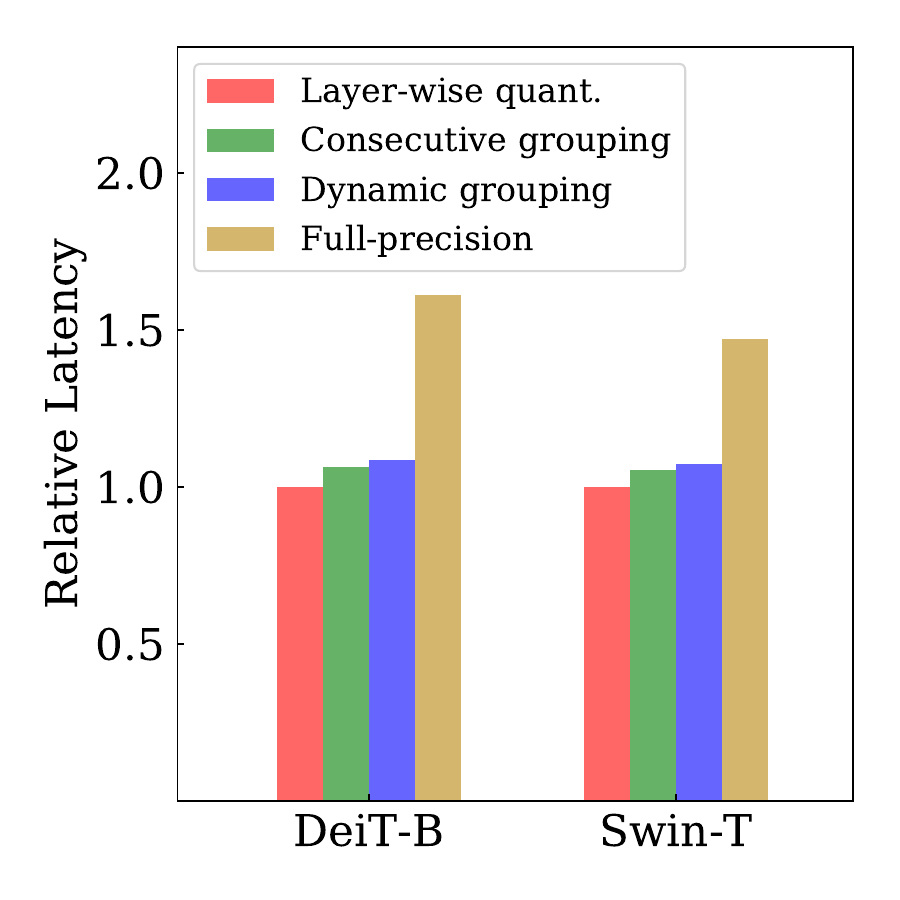}
          \vspace{-7mm}
          \caption{}
          \label{fig:rebuttal_1}
      \end{subfigure}
      \begin{subfigure}[b]{0.45\linewidth}
          \centering
          \includegraphics[width=1\linewidth]{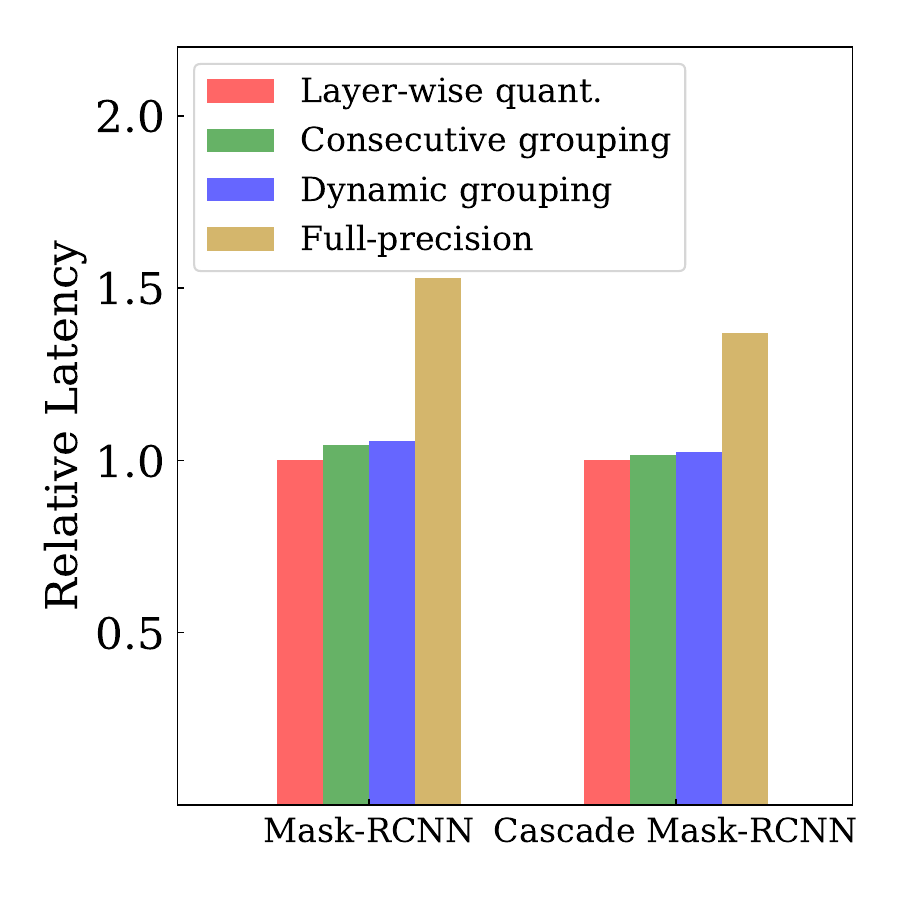}
          \vspace{-7mm}
          \caption{}
          \label{fig:rebuttal_2}
      \end{subfigure}
      \vspace{-3mm}
      \captionsetup{font={footnotesize}}
      \caption{Run-time latency for the tasks of (a) image classification and (b) object detection/instance segmentation using NVIDIA RTX 3090. For group quantization, we use a group size of 8 for all layers. For Mask-RCNN models, we use Swin-T as the backbone.}
      \label{fig:rebuttal}
\end{center}
\vspace{-6mm}
\end{figure}

\section*{C. Latency on practical devices}
To further validate the efficiency of IGQ-ViT, we conduct a simulation in PyTorch to compare the latencies between prior group quantization techniques~\cite{bondarenko2021understanding, dai2021vs, shen2020q} and ours. A key challenge is that most quantization methods exploit a fake quantization approach, following~\cite{li2021mqbench}, which mimics the quantization process by discretizing the network's weights and activations into a finite set of floating-point values, thereby serving a surrogate for the true quantization process. This approach is inappropriate for estimating the latency on real hardwares as it does not change the actual precision of the data, but merely introduces the concept of lower precision during calibration. Accordingly, we directly convert the data formats of weights and activations into 8-bit representations to measure the latency more accurately. Specifically, we simulate IGQ-ViT for linear operations using Eq.~(7), which requires low-bit matrix multiplication between weights and activations within each group, along with the summation of outputs for each group in full-precision. Since PyTorch does not support convolutional or linear layers that takes low-bit matrices as input, we have implemented their 8-bit counterparts. Note that we have implemented the group assignment algorithm~(\ie, Eq.~(5)) in full-precision.

We compare in Fig.~\ref{fig:rebuttal} the run-time latency of IGQ-ViT with its variants. We can see that IGQ-ViT introduces marginal overhead compared to layer-wise quantization, and consecutive grouping strategy, while achieving high quantization performances (See Table~5 in the main paper). This suggests that dynamic grouping of channels have a limited impact on actual latency.

\renewcommand\thetable{A}
\begin{table}[t]
  \setlength{\tabcolsep}{0.3em}
  \centering
  \bgroup
  \renewcommand{\arraystretch}{1.1}
  \setlength{\tabcolsep}{1.2pt}
  \caption{Results of quantizing a DETR model with a ResNet-50~\cite{he2016deep} backbone on COCO~\cite{lin2014microsoft}.}
  \vspace{-3mm}
  \label{tab:detr}
  \begin{tabular}{c|c|c}
  \hline
  \multicolumn{1}{c|}{\bf{Method}} & \bf{\begin{tabular}{@{}c@{}}\#bits\\(W/A)\end{tabular}} & \bf{\begin{tabular}{@{}c@{}}Box AP\\(Latency)\end{tabular}} \\ 
  \hline
  Full-precision & 32/32 & 42.0 \\
  \hline
  PTQ for ViT~\cite{liu2021post} & 6/6 & 40.5 \\
  \hline
  Ours (\#groups=8) & 6/6 & 41.1 \\
  Ours (\#groups=12) & 6/6 & {\bf{41.3}} \\
  \hline
  \end{tabular}
  \egroup
\end{table}

\section*{D. Application to DETR}
We show in Table~\ref{tab:detr} the results of quantizing a DETR model with a ResNet-50~\cite{he2016deep} backbone on COCO~\cite{lin2014microsoft}. To the best of our knowledge, PTQ for ViT~\cite{liu2021post} is the only PTQ method that provides quantization results for a DETR model under 6/6-bit setting. We can see that IGQ-ViT outperforms it by 0.8\% for a group size of 12.

\end{document}